\documentclass[conference]{IEEEtran}
\usepackage{times}

\usepackage[numbers]{natbib}
\usepackage{multicol}
\usepackage[bookmarks=true]{hyperref}
\usepackage{amsmath, amssymb}
\usepackage{graphicx}
\usepackage{xcolor}
\usepackage{xcolor}
\usepackage{booktabs}
\usepackage{listings}
\usepackage{xcolor}
\usepackage{booktabs}
\usepackage{multirow}

\begin{document}

\title{Semantically Structured Mixture-of-Experts for Compositional Robotic Manipulation}




%
\author{
\authorblockN{
Chengyu Deng\authorrefmark{2}\authorrefmark{1},
Guanqi Chen\authorrefmark{2}\authorrefmark{3}\authorrefmark{1},
Yizhou Chen\authorrefmark{2}, 
Zejia Liu\authorrefmark{2},
Zhiwen Ruan\authorrefmark{3},
Guanhua Chen\authorrefmark{3},
Jia Pan\authorrefmark{2}
}
\authorblockA{\authorrefmark{2}The University of Hong Kong \quad \authorrefmark{3}Southern University of Science and Technology}
\authorblockA{\authorrefmark{1}Equal contribution}
}

\maketitle

\begin{abstract}
Diffusion-based policies have established a new standard for precise robotic manipulation but face a critical scalability bottleneck: high-performance models are computationally expensive, while lightweight alternatives often fail to generalize across diverse multi-task environments. Mixture-of-Experts (MoE) architectures offer a promising path to efficiency by activating only a subset of parameters. However, existing MoE routing mechanisms typically rely on low-level noise or latent statistics, ignoring the compositional nature of manipulation tasks. This can fragment reusable behaviors across experts, limiting interpretability and transferability. We introduce Semantically Structured Mixture-of-Experts Diffusion Policy (SMoDP) for compositional robotic manipulation, a framework that grounds expert specialization in semantic task structure. SMoDP leverages a lightweight, inference-time skill predictor, supervised by offline annotations from Vision-Language Models (VLMs), to route action chunks to experts specialized for specific behavioral phases. To ensure robust assignment, we propose a dual contrastive alignment strategy that grounds multi-modal observations in language-defined skill semantics (Inter-modal) while enforcing routing consistency across visually distinct but functionally related behaviors (Intra-modal). Our approach outperforms representative diffusion and MoE-based baselines on multi-task benchmarks with significantly improved parameter efficiency and demonstrates effective compositional transfer to novel tasks through parameter-efficient fine-tuning. Project website: \url{https://deng-cy20.github.io/SMoDP/}

\end{abstract}

\IEEEpeerreviewmaketitle

\section{Introduction}
\label{section_intro}
Diffusion-based policies have recently emerged as a powerful paradigm for robotic manipulation~\cite{chi2023diffusion, prasad2024consistency, yu2024bikc, Ze2024DP3, chen2025svip}, achieving remarkable success in single-task scenarios through their ability to model multimodal action distributions and handle complex, high-dimensional control problems. By modeling action generation as conditional denoising, diffusion policies can represent multimodal action distributions and produce temporally coherent action chunks.



However, scaling these generative policies to general-purpose, multi-task settings presents a critical trilemma: we simultaneously desire (1) the precise control of diffusion policies, (2) the broad generalization of large-scale foundation models, and (3) the inference efficiency required for real-time deployment. Current approaches typically compromise on at least one front. Standard diffusion policies often struggle to generalize across diverse tasks without massive parameter scaling. Conversely, recent ``robot foundation models" such as RDT~\cite{liurdt} and Pi-series~\cite{black2026pi0,intelligence2025pi05} achieve breadth by scaling to hundreds of millions or billions of parameters. While effective, these massive architectures incur substantial computational overhead, making real-time deployment on resource-constrained robotic platforms impractical.

\begin{figure}[t]
  \centering
  \includegraphics[width=\linewidth]{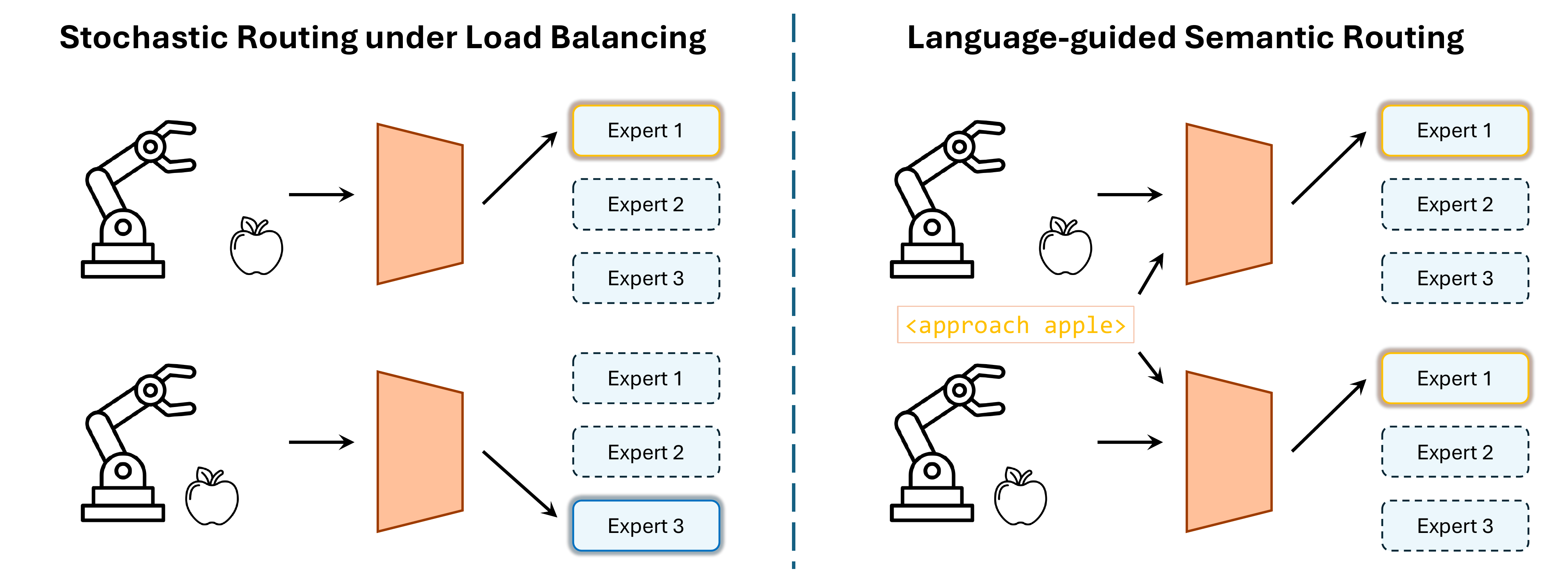} 
  \caption{Routing comparison between stochastic load-balanced routing (left) and our proposed language-guided, skill-aware semantic routing (right). Under a load-balancing objective, stochastic routing may assign different experts to generate similar or identical actions, leading to suboptimal expert assignments. In contrast, our language-guided semantic routing mechanism enables skill-aware expert assignment, where similar or identical skills are processed by the same experts.}
  \label{fig:intro}
\end{figure}


A principled solution to this efficiency-capacity trade-off is conditional computation, specifically through Mixture-of-Experts (MoE) architectures~\cite{reussefficient, huangmentor, cheng2025moe, liu2026factorizeddiffusionpolicy,wang2025sparse}. By activating only a sparse subset of parameters per inference step, MoEs can scale model capacity without a corresponding increase in compute. Recent work such as MoDE~\cite{reussefficient} has successfully integrated MoE into diffusion policies by routing computation based on diffusion noise levels. However, these methods primarily rely on low-level signals (e.g., noise or latent variance), crucially overlooking the semantic and temporal structure of manipulation tasks. Because noise-based routing lacks explicit temporal awareness, it may fail to provide consistent guidance at the transitions between distinct behavioral phases. This leads to ``routing confusion" at skill boundaries, where shared behaviors—such as ``grasping"—are fragmented across disparate experts (see Figure~\ref{fig:intro}), limiting both transferability and interpretability.


Recent work~\cite{jiangcr, omi2025loadbalancingmixtureexperts} has shown that introducing structured guidance into MoE routing can promote more coherent expert specialization and improve both performance and efficiency. These approaches demonstrate that routing decisions need not be driven purely by unconstrained correlations in the training data, but can benefit from additional structural signals. However, prior methods~\cite{liang2024skilldiffuser, wu2025discrete, mete2024quest, smith2025steer, hao2026abstracting} either use language or skill abstractions as an external interface for planning and skill orchestration, or discover reusable skills from latent or action-space structure, rather than explicitly using language-grounded skill semantics to organize expert routing. In contrast, we explore skill-conditioned routing in the context of multi-task imitation learning, where skills exhibit consistent semantic meaning across tasks and directly guide expert specialization inside the policy.

In this work, we aim to transfer knowledge from existing skills to semantically similar new skills, while minimizing the number of parameters and computational cost. To this end, we propose Semantically Structured Mixture-of-Experts Diffusion Policy (SMoDP), a novel framework that integrates skill-conditioned expert routing into multi-task diffusion policies. Our approach introduces two key innovations. First, we design a VLM-aided skill abstraction module to partition a long demonstration into a sequence of semantically meaningful skills. A skill predictor learns to predict the upcoming skill from a multimodal context, enabling skill-aware routing decisions. Second, we introduce a dual contrastive learning strategy to ensure effective skill-based expert specialization: inter-modal contrastive learning aligns the skill predictor's outputs with VLM-generated descriptions, bridging states with language-based skill semantics; intra-modal contrastive learning supervises the router to produce consistent routing distributions for similar skills, encouraging functionally related skills to activate overlapping expert subsets. This dual learning paradigm enables interpretable and parameter-efficient expert specialization by ensuring that the router leverages rich semantic information while maintaining routing consistency across similar skills. 
By aligning the policy's modular architecture with the semantic structure of the task, SMoDP achieves the highest performance among all evaluated methods on multi-task benchmarks. It not only improves parameter efficiency but also enables compositional transfer: by fine-tuning only the skill predictor and router while freezing expert weights, the model can tackle novel tasks by recombining previously learned skill modules.

In summary, our contributions are as follows:
\begin{enumerate}
    \item We introduce SMoDP, a semantically structured, skill-conditioned routing framework for diffusion policies that leverages language-based skill representations to guide expert selection, enabling more interpretable and structured expert specialization for multi-task robotic manipulation.
    \item We develop an offline VLM-aided skill abstraction module that automatically annotates demonstrations with open-vocabulary verb--noun skill segments, providing semantic supervision without manual labels or inference-time VLM calls.
    \item We propose a dual contrastive learning strategy that aligns skill prediction with language semantics and enforces routing consistency across related skills, encouraging skill-aligned expert specialization.
    \item We empirically validate SMoDP on both simulation and real-world benchmarks, demonstrating improved data efficiency, parameter efficiency, and compositional transfer through parameter-efficient fine-tuning.
\end{enumerate}

\section{Related Work}

\subsection{Diffusion Models for Robotic Manipulation}

Diffusion models~\cite{song2019generative, ho2020denoising, karras2022elucidating}, originally developed for generative modeling in vision and language, have recently been adapted for learning robot policies from demonstrations. In imitation learning, Diffusion Policy~\cite{chi2023diffusion} formulates action generation as a conditional denoising process, enabling multimodal and temporally coherent action sequences and achieving strong performance in single-task manipulation. Subsequent work has explored more efficient or expressive variants, including flow-matching policies~\cite{zhang2025flow} and consistency policies~\cite{prasad2024consistency}, extensions to 3D spatial representations~\cite{Ze2024DP3, ke20253d}, improved temporal modeling with dynamics constraints~\cite{liu2024diff}, and applications to long-horizon or multi-stage manipulation~\cite{chen2025svip, yu2024bikc, xu2025bikcplus}.

Despite these advances, generalizing diffusion policies to diverse multi-task settings remains a key challenge in robotics. One line of work addresses this challenge by scaling up model capacity and training on massive, heterogeneous datasets~\cite{liurdt, black2026pi0, barreiros2025careful, intelligence2025pi05, o2024open, khazatsky2024droid}. While effective, such approaches rely on extensive data collection and large models, limiting practicality for many robotic platforms. As a result, an alternative line of research focuses on improving generalization and efficiency under data-limited conditions. 
Some methods learn structured latent spaces or skill abstractions to capture shared structure across tasks~\cite{liang2024skilldiffuser, wu2025discrete, mete2024quest}. Closely related, STEER~\cite{smith2025steer} explores language-based skill orchestration with a VLM in the control loop, while SMP~\cite{hao2026abstracting} abstracts manipulation through action-space primitives. Other approaches introduce architectural innovations, such as Mixture-of-Experts (MoE) diffusion policies~\cite{reussefficient, wang2025sparse}, which reduce computation through sparse expert activation and enable partial parameter reuse across tasks.

However, existing approaches for multi-task diffusion policies largely rely on latent representations learned from data to capture shared structure across tasks. While effective in practice, such latent abstractions do not explicitly encode the human understanding of manipulation tasks, making it difficult to ensure consistent reuse of shared behaviors across tasks—particularly under limited data or novel task compositions. This motivates the incorporation of explicit structural inductive biases that reflect the compositional nature of manipulation skills.

\subsection{Mixture-of-Experts}

Mixture-of-Experts (MoE)~\cite{jacobs1991adaptive, jordan1994hierarchical} architectures enable efficient scaling by sparsely activating subsets of parameters conditioned on the input. MoE has seen renewed interest in large language models~\cite{shazeer2017outrageously, fedus2022switch, lepikhin2021gshard}, where sparse activation allows models with extremely large capacity to be trained and deployed efficiently. Recently, MoE architectures~\cite{huangmentor, cheng2025moe, liu2026factorizeddiffusionpolicy, wang2025sparse, reussefficient} have been adopted for robotic manipulation to address the computational and data efficiency challenges of multi-task policy learning. For example, Sparse-DP~\cite{wang2025sparse} explores task-specific expert specialization, while MoDE~\cite{reussefficient} integrates MoE into diffusion policies via noise-conditioned routing, significantly reducing inference cost while maintaining competitive multi-task performance.

Despite these advances, existing MoE-based imitation policies predominantly rely on data-driven routing mechanisms~\cite{fedus2022switch, lepikhin2021gshard, riquelme2021scaling}, where expert assignments are induced implicitly from training dynamics rather than being aligned with semantically meaningful behavioral structure. Recent work on structured routing has shown that introducing additional constraints can improve expert coherence and efficiency: CR-MoE~\cite{jiangcr} leverages consistency regularization to stabilize routing across similar inputs, and \citet{omi2025loadbalancingmixtureexperts} propose similarity-preserving objectives that reduce expert fragmentation and improve load balancing. However, these approaches derive structure from latent similarity rather than from task- or skill-level semantics.

In contrast, we introduce skill-conditioned routing tailored to robotic manipulation scenarios, explicitly conditioning expert selection on language-based skill semantics. By grounding routing decisions in reusable manipulation skills, our approach promotes consistent expert reuse across tasks and task compositions, enabling more interpretable and structured expert specialization in multi-task imitation learning.

\section{Method}

We present \emph{Semantically Structured Mixture-of-Experts Diffusion Policy} (SMoDP), a skill-conditioned routing framework for multi-task diffusion policies. Existing MoE routing mechanisms~\cite{reussefficient,wang2025sparse} primarily rely on statistical patterns without exploiting task structure or skill composition, leading to suboptimal expert specialization where similar skills across different tasks are processed by different experts. 

\begin{figure*}[t]
        \centering
        \includegraphics[width=\linewidth]{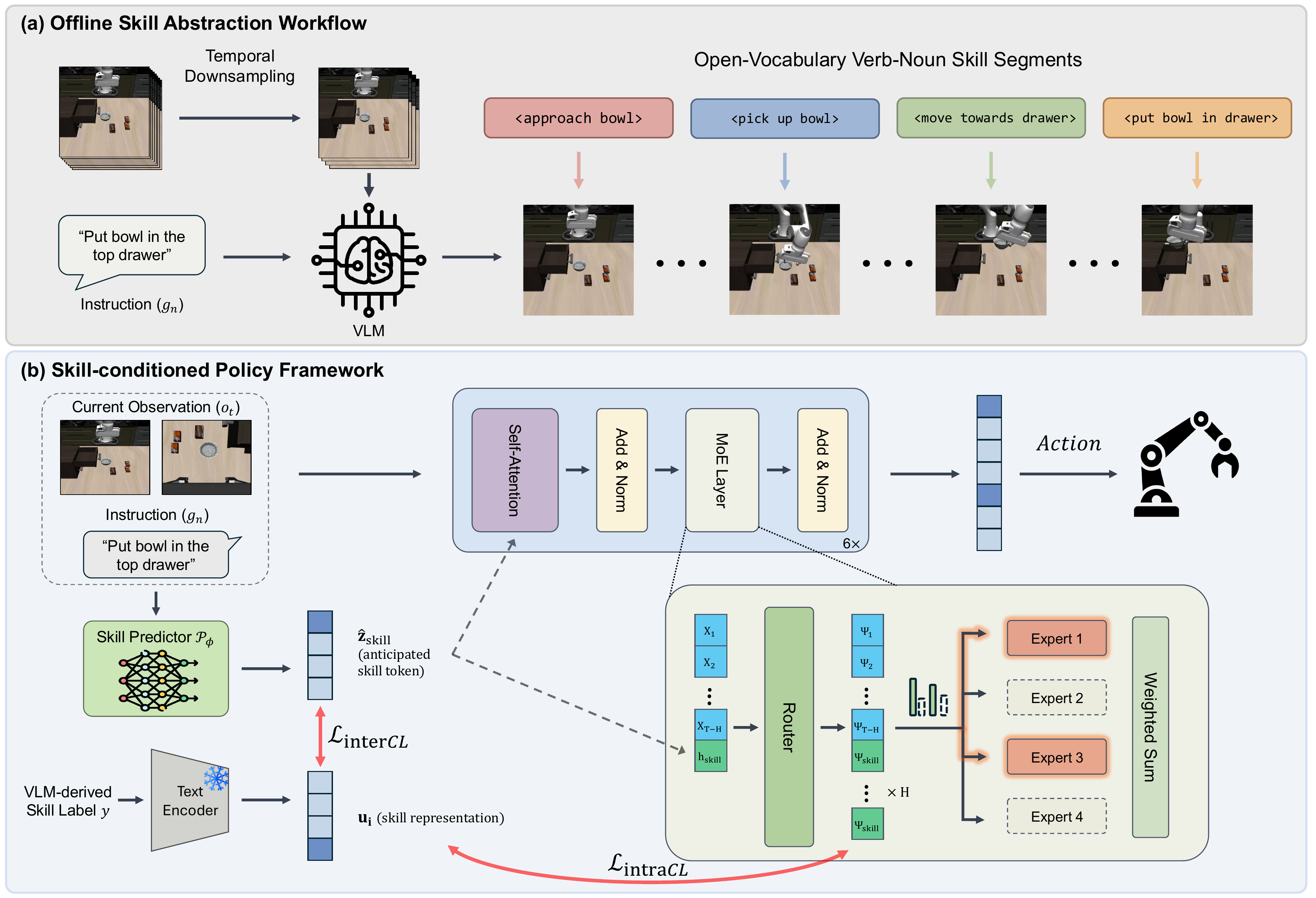} 
        \caption{Overview of SMoDP: (a) \textbf{Offline Skill Abstraction}: A workflow that automatically annotates demonstrations with open-vocabulary verb--noun skills, eliminating the need for manual labeling. (b) \textbf{Skill-Conditioned Diffusion MoE Policy}:  A framework that leverages a lightweight skill predictor to anticipate the upcoming skill from multimodal context, then performs chunk-consistent expert routing with dual semantic contrastive alignment—ensuring that functionally related skills activate overlapping experts while enabling skill-aware control.}
        \label{fig:main}
\end{figure*}

SMoDP overcomes this limitation through two key innovations: (i) \textbf{offline VLM-based skill abstraction} that automatically annotates demonstrations with open-vocabulary verb--noun skills, avoiding manual skill labeling; (ii) a \textbf{skill-conditioned diffusion MoE policy} that employs a lightweight online skill predictor to infer skill embeddings from multimodal context and performs chunk-consistent expert routing with dual semantic contrastive alignment, ensuring functionally related skills activate overlapping expert subsets while enabling skill-aware control. The overview of SMoDP is shown in Figure~\ref{fig:main}.

We consider multi-task imitation learning over a set of tasks $\mathcal{T}=\{\mathcal{T}_1,\dots,\mathcal{T}_M\}$ specified by language instructions $g$ (e.g., ``Pick up the red apple''). Given demonstrations $\mathcal{D}$, our goal is to learn a single visuomotor policy $\pi_\theta(\bar{\mathbf{a}}_{n,t}\mid \mathbf{o}_{n,t},g_n)$ that maps the current observation $\mathbf{o}_{n,t}$ and instruction $g_n$ to an $H$-step action chunk $\bar{\mathbf{a}}_{n,t}=[\mathbf{a}_{n,t},\dots,\mathbf{a}_{n,t+H-1}]$. We train by maximizing the demonstration log-likelihood:
\begin{equation}
    \mathcal{L}_{\mathrm{IL}} =
    \mathbb{E}_{(\tau_n,g_n)\sim \mathcal{D}}\;
    \mathbb{E}_{t}\Big[\log \pi_\theta(\bar{\mathbf{a}}_{n,t} \mid \mathbf{o}_{n,t}, g_n)\Big].
\end{equation}


\subsection{Offline Semantic Skill Abstraction}
\label{sec:vlm_skills}

A key challenge in implementing skill-conditioned routing is the lack of fine-grained skill annotations in existing robotic datasets. Manual labeling is prohibitively expensive and requires domain expertise. We address this by leveraging Vision-Language Models (VLMs) to automatically construct offline skill annotations from unstructured demonstrations, exploiting their zero-shot reasoning and visual understanding capabilities. These annotations are used only during training to supervise skill prediction and routing regularization; the VLM is not invoked during policy inference.

For a trajectory $\tau_n$ with task instruction $g_n$, we use a VLM to decompose $\tau_n$ into a sequence of $K_n$ semantic skill segments $\mathcal{S}_n=\{s_{n,1},\dots,s_{n,K_n}\}$ with temporal boundaries $\{[t^{(j)}_{\mathrm{start}},t^{(j)}_{\mathrm{end}}]\}_{j=1}^{K_n}$. Each skill is represented as a verb--noun pair (e.g., $s_{n,j}=\langle \texttt{pick up},\texttt{red cup}\rangle$), yielding open-vocabulary skill supervision without a predefined label set. Our annotation pipeline consists of three stages:

\begin{itemize}
\item \textbf{Video Downsampling.}
Raw observation sequences in robotic datasets are typically high-frequency, containing redundant frames that contribute little to high-level reasoning while increasing the VLM token budget. We convert the image sequence in $\tau_n$ into a video format $\mathcal{V}_n$ using a temporal downsampling factor $\lambda$: $\mathcal{V}_n = \mathrm{Downsample}(\tau_n, \lambda)$. This operation mitigates potential hallucinations by filtering out non-essential temporal jitter while preserving critical state transitions needed for accurate action recognition.

\item \textbf{Semantic Skill Extraction.}
The downsampled video $\mathcal{V}_n$ and task instruction $g_n$ are jointly fed into the VLM. We employ a structured prompting strategy that enforces a canonical output format, tasking the VLM with analyzing the trajectory and distilling the behavior into a sequence of atomic skills $\mathcal{S}_n$. Crucially, we constrain the output space to \textit{verb--noun} pairs (e.g., $s_{n,j} = \langle \texttt{pick up}, \texttt{red cup} \rangle$), which serves as a structured semantic bottleneck that enhances interpretability and compositionality.

\item \textbf{Temporal Partitioning.}
For each identified skill $s_{n,j}\in\mathcal{S}_n$, the VLM is prompted to estimate the corresponding temporal boundary $[t^{(j)}_{\mathrm{start}}, t^{(j)}_{\mathrm{end}}]$ within the original trajectory $\tau_n$. 
\end{itemize}

Through this process, we transform $\mathcal{D}$ into a semantically labeled dataset $\mathcal{D}_{\mathrm{sem}}$, where each timestep $t$ is assigned a skill label $y_{n,t}=s_{n,j}$ if $t\in[t^{(j)}_{\mathrm{start}},t^{(j)}_{\mathrm{end}}]$. These labels supervise the online skill predictor and provide semantic structure for MoE routing. The detailed prompts and examples are provided in the supplementary material.

\subsection{Skill-Conditioned Diffusion MoE Policy}
\label{sec:diffusion_moe}

Building upon the diffusion policy formulation~\cite{chi2023diffusion}, SMoDP integrates skill semantics into expert selection through two key components: a lightweight skill predictor that efficiently infers skill phases from multimodal context at inference time, and a chunk-consistent MoE architecture that uses the predicted skill embedding to guide expert routing.

\paragraph{Diffusion Policy Over Action Chunks}
Let $\mathbf{x}_0=\bar{\mathbf{a}}_{n,t}$ denote the demonstrated action chunk. Following the standard diffusion formulation, we perturb $x_0$ with Gaussian noise at diffusion step $s \in \{1, \ldots, S\}$:
\begin{equation}
    \mathbf{x}_s=\alpha_s \mathbf{x}_0+\sigma_s \boldsymbol{\epsilon},\quad \boldsymbol{\epsilon}\sim\mathcal{N}(\mathbf{0},\mathbf{I}),
\end{equation}
where $\{\alpha_s,\sigma_s\}$ follow a predefined noise schedule. We learn a conditional denoiser $f_\theta$ that predicts $\boldsymbol{\epsilon}$ from the noisy action chunk $\mathbf{x}_s$, observation $\mathbf{o}_{n,t}$, task instruction $g_n$, and diffusion step $s$ using the denoising objective:
\begin{equation}
    \mathcal{L}_{\mathrm{diff}}=\mathbb{E}_{\mathbf{x}_0,s,\boldsymbol{\epsilon}}\big[\|\boldsymbol{\epsilon}-f_\theta(\mathbf{x}_s,\mathbf{o}_{n,t},g_n,s)\|_2^2\big].
    \label{eq:diff}
\end{equation}

\paragraph{Lightweight Skill Predictor}
While VLM-generated skill annotations provide rich supervision during training, invoking a large VLM at inference time is impractical for real-time robotic control. We therefore train a lightweight skill predictor $\mathcal{P}_\phi$ that infers a compact semantic skill embedding from the policy's multimodal context.

Let the context token sequence be $\mathbf{C}=[\mathbf{e}_{\mathrm{obs}}; \mathbf{e}_{\mathrm{goal}}]\in\mathbb{R}^{T_{\mathrm{ctx}}\times d}$, where $\mathbf{e}_{\mathrm{obs}}$ denotes visual tokens from the observation and $\mathbf{e}_{\mathrm{goal}}$ denotes language tokens from the task instruction. We introduce a learnable query $\mathbf{q}_{\mathrm{skill}}\in\mathbb{R}^{d}$ to extract skill-relevant information via cross-attention, followed by a feed-forward projection into the frozen text-embedding space $\mathbb{R}^{d_u}$:
\begin{equation}
    \hat{\mathbf{z}}_{\mathrm{skill}} =
    \mathrm{FFN}\!\Big(
    \mathrm{CrossAttn}(\mathbf{q}_{\mathrm{skill}}, \mathbf{C}, \mathbf{C})\Big).
\end{equation}
By predicting a continuous embedding in the same space as the frozen text encoder, 
the skill predictor can be supervised by language-derived skill labels. This query-based design is computationally lightweight, employing only a single cross-attention layer and a compact projection module, while producing a semantic skill representation used for subsequent routing.

\paragraph{Skill-Conditioned MoE Denoising}
We implement the denoiser $f_\theta$ as a Diffusion Transformer in which the feed-forward networks are replaced by MoE blocks. Let $\mathbf{H}^0\in\mathbb{R}^{T\times d}$ denote the token sequence obtained by concatenating context tokens with $H$ action tokens representing the noisy action chunk $\mathbf{x}_s$, where $T=T_{\mathrm{ctx}}+H$. We condition the transformer on the predicted skill embedding through a projected conditioning vector $\mathbf{c}=\mathbf{W}_c \hat{\mathbf{z}}_{\mathrm{skill}}\in\mathbb{R}^{d}$, which is broadcast to all tokens. At layer $l$, the update follows:
\begin{equation}
    \label{eq:overall}
    \begin{aligned}
    \mathbf{X}^l &= \mathbf{H}^{l-1} + \mathrm{Attn}^l\!\Big(\mathbf{H}^{l-1} + \mathbf{1}\mathbf{c}^\top\Big),\\
    \mathbf{H}^l &= \mathbf{X}^{l} + \mathrm{MoE}^l\!\Big(\mathrm{LN}(\mathbf{X}^{l});\,\hat{\mathbf{z}}_{\mathrm{skill}}\Big),
    \end{aligned}
\end{equation}
where $\mathbf{1}\in\mathbb{R}^{T\times 1}$ denotes an all-ones vector for broadcasting, $\mathrm{LN}(\cdot)$ denotes layer normalization, and $\mathrm{Attn}^l(\cdot)$ denotes the self-attention module.

\paragraph{Chunk-Consistent Skill-Based Routing}
The MoE layer incorporates the predicted skill embedding $\hat{\mathbf{z}}_{\mathrm{skill}}$ into expert routing to encourage chunk-level consistency. Since each denoising step predicts a short action chunk, allowing different action tokens within the same chunk to select unrelated experts may introduce inconsistent local control. We therefore route all action tokens in the chunk using the same skill-conditioned router token, while keeping context-token routing token-specific.

Given token features $\mathbf{X}^l\in\mathbb{R}^{T\times d}$ at layer $l$, where the last $H$ tokens correspond to the action chunk, we compute a skill token $\mathbf{h}_{\mathrm{skill}}=\mathbf{W}_z\hat{\mathbf{z}}_{\mathrm{skill}}\in\mathbb{R}^{d}$ and form the router input by replacing the action-token segment with this single skill token:
\begin{equation}
\tilde{\mathbf{H}}^{l}=\big[\mathbf{X}^{l}_{1:T-H},\ \mathbf{h}_{\mathrm{skill}}\big]\in\mathbb{R}^{(T-H+1)\times d}.
\end{equation}
Then we employ the standard multi-layer perceptron (MLP) as the router to predict the routing logits for each token:
\begin{equation}
\tilde{\boldsymbol{\Psi}}^{l}=\mathrm{MLP}(\tilde{\mathbf{H}}^{l})\in\mathbb{R}^{(T-H+1)\times E},
\end{equation}
where $E$ is the number of experts. To obtain chunk-level consistency, we broadcast the routing logits of the skill token to all $H$ action tokens. Concretely, we define full-sequence logits $\boldsymbol{\Psi}^{l}\in\mathbb{R}^{T\times E}$ by:
\begin{equation}
\boldsymbol{\Psi}^{l}_{1:T-H}\leftarrow \tilde{\boldsymbol{\Psi}}^{l}_{1:T-H},\quad
\boldsymbol{\Psi}^{l}_{T-H+1:T}\leftarrow \tilde{\boldsymbol{\Psi}}^{l}_{T-H+1}.
\end{equation}
We apply temperature-scaled softmax followed by top-$k$ sparsification to obtain expert gates:
\begin{equation}
\mathbf{G}^{l}=\mathrm{top}\text{-}k\!\left(\mathrm{softmax}\!\left(\boldsymbol{\Psi}^{l}/\tau_r\right)\right),
\end{equation}
where $\tau_r$ is the router temperature and $\mathbf{G}^{l}$ keeps only the top-$k$ experts for each token. Given expert networks $\{{E}^{l}_e(\cdot)\}_{e=1}^{E}$, the MoE output for token $t$ is computed as a gated mixture:
\begin{equation}
\mathrm{MoE}^{l}(\mathbf{h}_t)=\sum_{e=1}^{E} G^{l}_{t,e}\, {E}^{l}_e(\mathbf{h}_t).
\end{equation}
This \emph{chunk-consistent, skill-based routing} reduces within-chunk expert switching and encourages skill-aligned expert specialization across tasks.

\subsection{Dual Semantic Contrastive Alignment}
\label{sec:contrastive_learning}
To encourage the skill predictor and MoE router to leverage the semantic structure provided by VLM annotations, we introduce a \emph{dual semantic contrastive alignment} strategy.
This approach consists of two complementary objectives: (i) \emph{Inter-Modal Contrastive Learning (InterCL)} that aligns the predicted skill embeddings with frozen language-encoder embeddings, bridging states with language-based skill semantics, and (ii) \emph{Intra-Modal Contrastive Learning (IntraCL)} that supervises the router to produce consistent routing distributions for semantically similar skills, encouraging functionally related skills to activate overlapping expert subsets. 
InterCL shapes the predicted skill representation, whereas IntraCL directly regularizes the router behavior; thus, the former provides language-grounded skill semantics, while the latter encourages such semantics to be reflected in expert selection. 
Together with the diffusion denoising loss $\mathcal{L}_{\mathrm{diff}}$ defined in Eq.~(\ref{eq:diff}) and the standard MoE load-balancing regularizer $\mathcal{L}_{\mathrm{lb}}$~\cite{fedus2022switch}, this dual alignment encourages interpretable and parameter-efficient expert specialization.

\paragraph{Language-Derived Continuous Weights}
Inspired by CWCL~\cite{srinivasa2023cwcl}, we also adopt continuous similarity weights for skill alignment. For a minibatch of size $B$, let $\mathbf{u}_i=\mathrm{norm}(\mathrm{TextEnc}(y_i))\in\mathbb{R}^{d_u}$ denote the normalized text embedding of the VLM-generated skill label $y_i$ from a frozen text encoder. We construct a non-negative semantic affinity matrix $\mathbf{W}\in\mathbb{R}^{B\times B}$ from pairwise text-embedding similarity:
\begin{equation}
d_{ij}=\max \left( \langle \mathbf{u}_i,\mathbf{u}_j\rangle, 0 \right),\quad d_{ii}=0.
\end{equation}
To reduce noise from spurious positive pairs, we retain only the top-$m$ most similar samples for each anchor $i$, where $m=\lceil \rho (B-1)\rceil$ and $\rho\in(0,1]$ denotes a hyperparameter controlling the keep ratio. Let $\mathrm{top}\text{-}m(d_i)$ denote the index set of the $m$ largest entries in row $d_i=\{d_{ij}\}_{j=1}^{B}$. We define the weight matrix entries as:
\begin{equation}
w_{ij}=
\begin{cases}
d_{ij}, & j\in \mathrm{top}\text{-}m(d_{i}), \\
0, & \text{otherwise}.
\end{cases}
\end{equation}

\paragraph{Inter-Modal Contrastive Learning (InterCL)}
Let $\mathbf{q}_i=\mathrm{norm}(\hat{\mathbf{z}}_{\mathrm{skill},i})\in\mathbb{R}^{d_u}$ be the normalized predicted skill embedding for sample $i$. InterCL bridges the gap between the learned skill predictor and the language-based semantic space by aligning $\mathbf{q}_i$ to frozen text embeddings via a weighted contrastive loss. We compute inter-modal similarity logits:
\begin{equation}
\ell_{ij}=\langle \mathbf{q}_i,\mathbf{u}_j\rangle/\tau_c,
\end{equation}
where $\tau_c$ is a temperature parameter. The InterCL loss is then formulated as:
\begin{equation}
\begin{aligned}
\mathcal{L}_{\mathrm{interCL}}
&=
-\frac{1}{B}\sum_{i=1}^{B}
\frac{1}{\sum_{j=1}^{B} w_{ij}}
\sum_{j=1}^{B} w_{ij}\,
\log
\frac{\exp(\ell_{ij})}{\sum_{k=1}^{B}\exp(\ell_{ik})}.
\end{aligned}
\end{equation}
Since $\mathbf{u}$ is produced by a pre-trained language encoder, this alignment encourages $\hat{\mathbf{z}}_{\mathrm{skill}}$ to inherit compositional structure and generalize by semantic proximity rather than memorizing a closed set of skill IDs.

\paragraph{Intra-Modal Contrastive Learning (IntraCL)}
While InterCL aligns predicted skill embeddings with language semantics, it does not directly constrain the router behavior. To encourage that semantically similar skills induce similar routing patterns, we introduce IntraCL, which operates on the routing logits themselves.

Let $\boldsymbol{\Psi}^{l}_i\in\mathbb{R}^{T\times E}$ be the broadcasted routing logits at MoE layer $l$ for sample $i$, and let $\mathcal{L}\subseteq\{1,\dots,L\}$ denote the set of MoE layers where we apply routing alignment. We flatten and normalize the routing logits into a router feature vector:
\begin{equation}
\mathbf{r}^{l}_i=\mathrm{norm}\!\left(\mathrm{vec}(\boldsymbol{\Psi}^{l}_i)\right)\in\mathbb{R}^{TE}.
\end{equation}
We then compute intra-modal similarity logits:
\begin{equation}
\ell^{l}_{ij}=\langle \mathbf{r}^{l}_i,\mathbf{r}^{l}_j\rangle/\tau_c.
\end{equation}
The IntraCL loss encourages routing consistency across semantically similar skills:
\begin{equation}
\begin{aligned}
\mathcal{L}_{\mathrm{intraCL}}
&=
\frac{1}{|\mathcal{L}|}\sum_{l\in\mathcal{L}}
\Bigg[
-\frac{1}{B}\sum_{i=1}^{B}
\frac{\sum_{j=1}^{B} w_{ij}\log p^l_{ij}}{\sum_{j=1}^{B} w_{ij}}
\Bigg],\\
\text{where}\quad
p^l_{ij} &= \frac{\exp(\ell^{\,l}_{ij})}{\sum_{k=1}^{B}\exp(\ell^{\,l}_{ik})}.
\end{aligned}
\end{equation}

By leveraging the language-derived weights $w_{ij}$, IntraCL acts as a topology-preserving constraint that compels the router to respect semantic distances: skills that are close in language space are encouraged to activate similar expert subsets. This regularization enhances expert specialization by preventing arbitrary partitioning across functionally related behaviors.

\paragraph{Overall Training Objective}
We combine the diffusion denoising loss, the standard MoE load-balancing regularizer, and the two semantic alignment terms:
\begin{equation}
\mathcal{L}
=
\mathcal{L}_{\mathrm{diff}}
+\lambda_{\mathrm{lb}}\mathcal{L}_{\mathrm{lb}}
+\lambda_{\mathrm{inter}}\mathcal{L}_{\mathrm{interCL}}
+\lambda_{\mathrm{intra}}\mathcal{L}_{\mathrm{intraCL}},
\end{equation}
where $\lambda_{\mathrm{lb}}$, $\lambda_{\mathrm{inter}}$, and $\lambda_{\mathrm{intra}}$ balance the load-balancing, InterCL, and IntraCL losses, respectively. This objective trains the model to denoise actions effectively while aligning both its skill representations and routing behavior with meaningful semantic structure derived from language.

\section{Experiments}
\label{sec:experiments}
In this section, we evaluate SMoDP in both simulated and real-world robotic manipulation settings to assess whether skill-conditioned expert routing improves multi-task learning, data efficiency, compositional transfer, and real-world execution. Using the LIBERO benchmark and a real bimanual ALOHA system, we compare SMoDP against representative diffusion, MoE-based, and skill-structured imitation learning baselines, together with targeted ablations.
Our experiments are designed to validate four core claims: (i) skill-aware routing improves multi-task performance, (ii) language-grounded skill structure improves data utilization in low-data regimes, (iii) frozen experts can be reused for parameter-efficient few-shot compositional transfer, and (iv) semantic routing improves multi-task learning ability on a real robot.

\subsection{Experimental Setup}
\label{sec:exp_setup}

\paragraph{Simulation Benchmark}
\begin{figure}[t]
    \includegraphics[width=\linewidth]{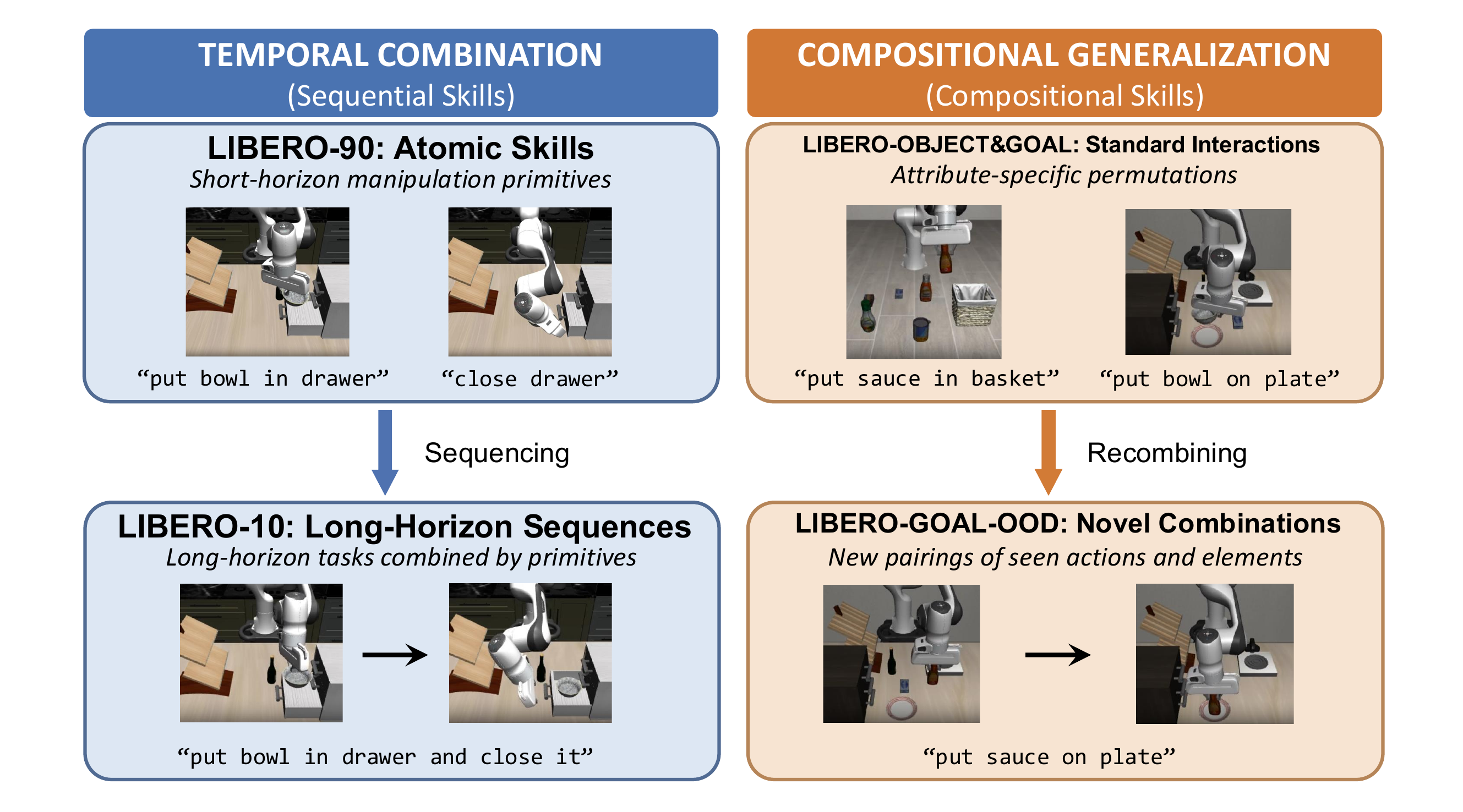}
    \caption{Overview of tasks in 4 LIBERO simulation task suites.}
    \label{fig_simulation_intro}
\end{figure}
We evaluate our method on the \textsc{LIBERO} suite~\cite{liu2023libero}, a standard multi-task manipulation benchmark with diverse object interactions and long-horizon compositions.
We use the following task suites:
(i) \textsc{LIBERO-90}, containing 90 manipulation tasks;
(ii) \textsc{LIBERO-10}, consisting of 10 long-horizon tasks derived from \textsc{LIBERO-90};
(iii) \textsc{LIBERO-OBJECT} and \textsc{LIBERO-GOAL}, designed to stress object-centric and goal-conditioned generalization, respectively; and
(iv) \textsc{LIBERO-GOAL-OOD}~\cite{li2025taskreconstructionextrapolationpi0}, a skill-level split constructed from \textsc{LIBERO-OBJECT/GOAL} to test compositional generalization to unseen skill combinations (e.g., ``put bowl on stove'' + ``put bottle on cabinet'' $\Rightarrow$ ``put bottle on stove'').
For task suites \textsc{LIBERO-90/10/OBJECT/GOAL}, 50 demonstrations are provided per task.
For \textsc{LIBERO-GOAL-OOD}, we manually collect 10 demonstrations per task and sub-sample them for few-shot adaptation (1/5/10 demos). Figure~\ref{fig_simulation_intro} shows some examples of these 4 task suites.
Each episode contains two RGB views, including a fixed front camera and a wrist-mounted camera. All methods are trained and evaluated using the same observations, demonstrations, and rollout protocol.

\begin{figure}
    \includegraphics[width=\linewidth]{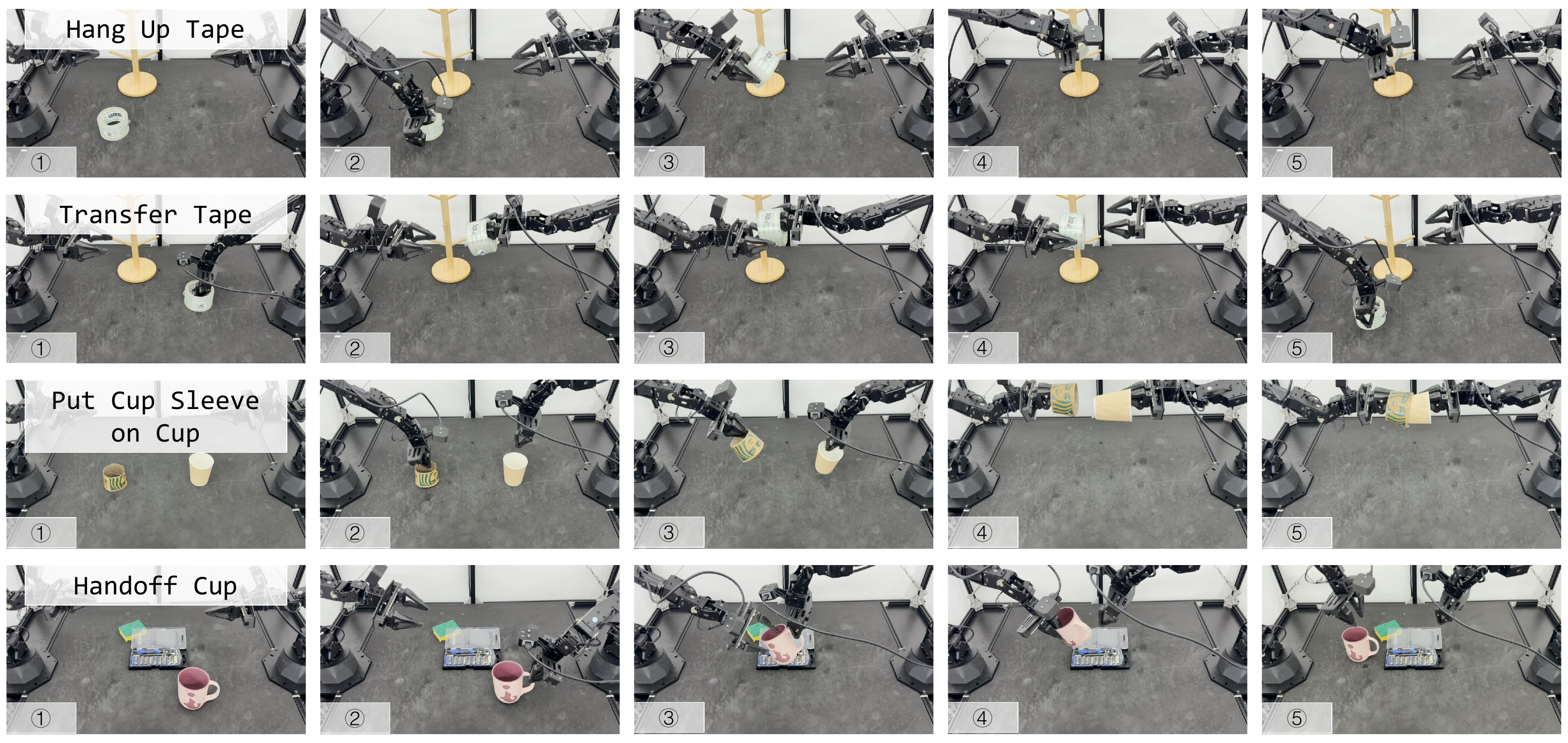}
    \caption{Real-world dual-arm ALOHA setup and task illustrations.} \label{fig:aloha}
\end{figure}

\paragraph{Real-world Benchmark}
We additionally validate our approach on a real ALOHA system for bimanual multi-task manipulation. 
Illustrations and real-world task setups are shown in Figure~\ref{fig:aloha}.
Our setup uses a 7-DoF master--puppet teleoperation system with four RGB cameras, including two wrist-mounted cameras and two external cameras from the main front view and the top-down view.
Unlike the single-arm simulation benchmark, the real-world ALOHA tasks require bimanual coordination and frequent hand--object occlusions; therefore, we provide the policy with a short visual history consisting of the most recent three image frames from all four cameras.
We collect 20 demonstrations per task for four real-world training tasks:
(i) \emph{Hang Up Tape}---pick up a tape roll with the left arm and hang it onto a shelf hook;
(ii) \emph{Transfer Tape}---pick up a tape roll with the right arm, hand it over to the left arm, and place it on the table;
(iii) \emph{Put Cup Sleeve on Cup}---pick up both cup and cup sleeve, and put cup sleeve on cup; and
(iv) \emph{Handoff Cup}---pick up cup and hand it off to the other hand.

\paragraph{Baselines}
We compare SMoDP against representative imitation learning baselines,
including diffusion-policy baselines with transformer or CNN backbones
(DP-T, DP-CNN)~\cite{chi2023diffusion}, a vector-quantized
transformer policy (QUEST)~\cite{mete2024quest}, and MoE-based diffusion
policies. Specifically, we include Sparse Diffusion Policy (SDP)~\cite{wang2025sparse},
which performs task-conditioned sparse expert activation, and
Mixture-of-Denoising Experts (MoDE)~\cite{reussefficient}, a transformer-based
diffusion policy that routes tokens to sparse denoising experts conditioned on
the diffusion noise level.
MoDE (Pretrain) denotes initializing MoDE from a checkpoint pre-trained on the large-scale OXE dataset~\cite{o2024open} before fine-tuning on the target \textsc{LIBERO} suite.
For transfer experiments, we additionally include: (i) MoDE with LoRA fine-tuning after pre-training on the source suite (MoDE+LoRA), and (ii) MoDE trained from scratch on the target suite (MoDE-scratch).

\paragraph{Implementation Details}
For offline VLM skill annotation, we use Qwen3-VL~\cite{bai2025qwen3vltechnicalreport} with a temporal downsampling factor of $\lambda=5$.
For the diffusion transformer, we use 6 transformer layers, 4 experts per MoE layer, and top-2 expert activation.
In simulation, we condition on the most recent observation frame from the two RGB views; in real-world ALOHA experiments, we use the most recent three frames from the four RGB cameras. In both settings, the policy predicts an action chunk with horizon $H=10$. For positive sample selection in our dual contrastive learning, we set the keep ratio to $\rho = 0.1$. We use Sentence-BERT~\cite{reimers-2019-sentence-bert} as the frozen text encoder to extract embeddings of VLM-generated skill annotations for dual semantic contrastive alignment. VLM annotation and contrastive alignment are used only during training; at inference time, SMoDP uses the learned lightweight skill predictor and does not call the VLM. In our implementation, SMoDP takes approximately 4 hours to train on a single NVIDIA RTX 4090. With 10 denoising steps, the average inference time is about 121 ms per action chunk, which is comparable to standard MoE diffusion policies.

\subsection{Multi-task Performance on \textsc{LIBERO-10} and \textsc{LIBERO-90}}

\begin{figure}[t]
    \includegraphics[width=\linewidth]{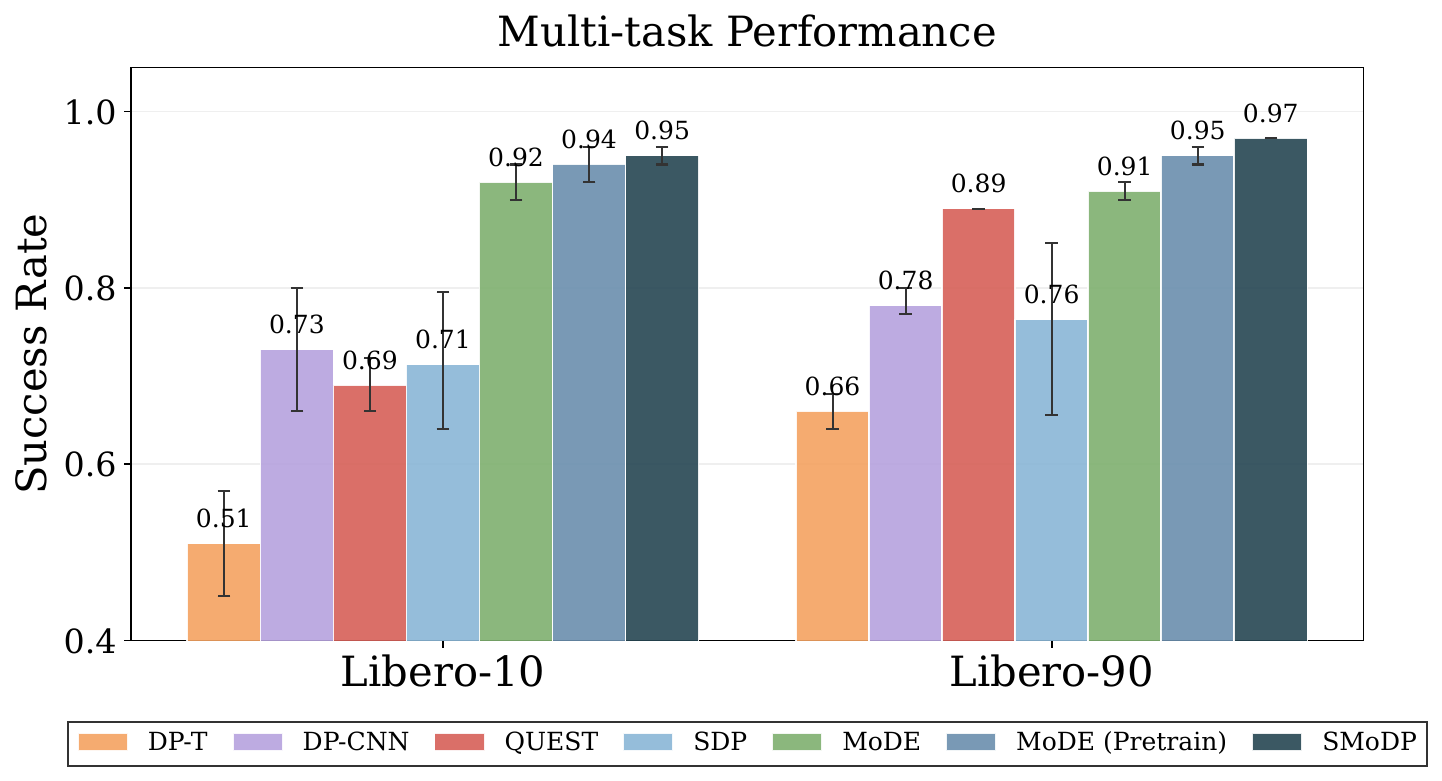}
    \caption{Average results for LIBERO-10 and LIBERO-90 averaged over 3 seeds with 20 rollouts per task.} \label{fig_libero90}
\end{figure}

We first evaluate multi-task imitation learning on \textsc{LIBERO-10} and \textsc{LIBERO-90} using 50 demonstrations per task. The results are shown in Figure~\ref{fig_libero90}. Overall, SMoDP achieves the highest average success rate on both benchmarks among the evaluated methods. Notably, SMoDP also outperforms MoDE (Pretrain), which spends more than 3 days pre-training on the large-scale OXE dataset. This suggests that language-grounded skill routing provides an effective inductive bias for sharing manipulation behaviors across tasks, allowing the policy to learn strong multi-task visuomotor representations without relying on external large-scale pre-training data. Compared with the fairer non-pretrained MoE baselines, the improvement indicates that the gain is not merely due to sparse expert capacity, but also comes from organizing expert activation around reusable skill semantics.

\begin{figure}[ht]
    \includegraphics[width=\linewidth]{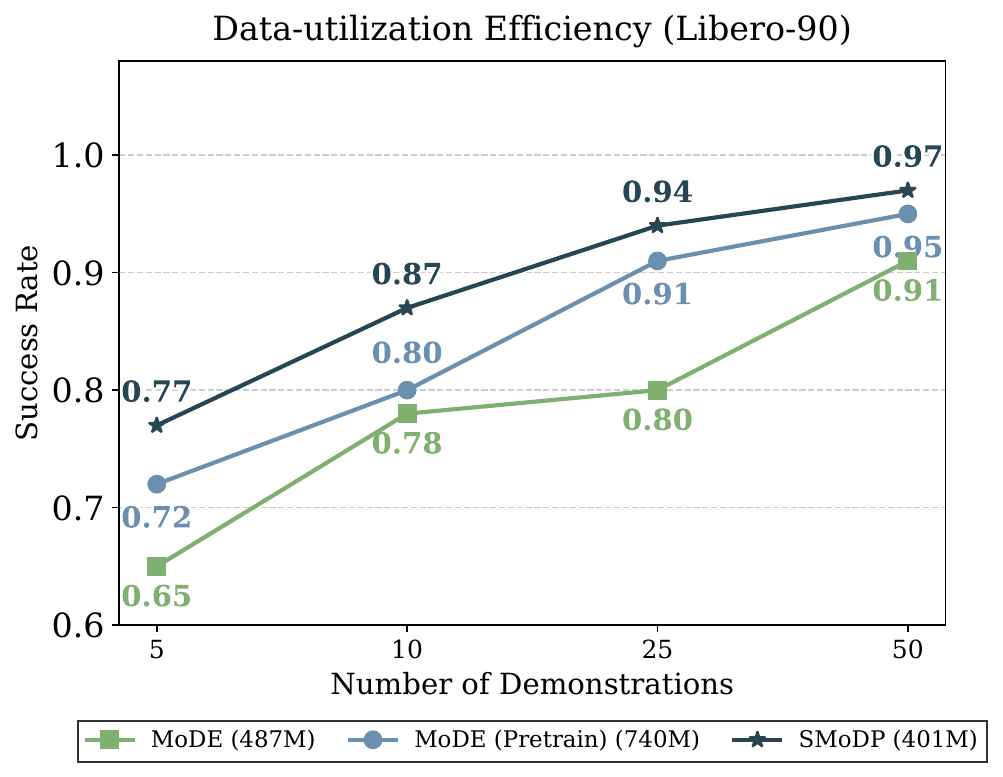}
    \caption{Average success rate on Libero-90 under different numbers of demonstrations for training.} \label{fig: data-util}
\end{figure}

\begin{table*}[ht]
    \centering
    \caption{Few-shot transfer success rate ($\uparrow$) on \textsc{LIBERO-10} and \textsc{LIBERO-GOAL-OOD}.}
    \label{tab:libero_comparison}
    \setlength{\tabcolsep}{8pt}
    \renewcommand{\arraystretch}{1.12}

    \begin{tabular}{lccccc ccc}
    \toprule
    \multirow{2}{*}{Method}
    & \multicolumn{2}{c}{Parameters (M)}
    & \multicolumn{3}{c}{\textsc{LIBERO-10}}
    & \multicolumn{3}{c}{\textsc{LIBERO-GOAL-OOD}} \\
    \cmidrule(lr){2-3}
    \cmidrule(lr){4-6}
    \cmidrule(lr){7-9}
    & Trainable & Total
    & 1 Demo & 5 Demos & 10 Demos
    & 1 Demo & 5 Demos & 10 Demos \\
    \midrule
    MoDE+LoRA
    & \textbf{13.6} & 487
    & 0.245 & 0.425 & 0.509
    & 0.319 & 0.410 & 0.635 \\

    MoDE-scratch
    & 487 & 487
    & 0.085 & 0.245 & 0.519
    & 0.170 & 0.129 & 0.495 \\

    \textbf{SMoDP (Ours)}
    & 13.7 & \textbf{401}
    & \textbf{0.520} & \textbf{0.765} & \textbf{0.840}
    & \textbf{0.320} & \textbf{0.590} & \textbf{0.660} \\
    \bottomrule
    \end{tabular}
\end{table*}
\label{sec:exp_multitask}

\begin{table*}[t]
\centering
\caption{Real-world robot experiment results on four tasks} \label{tab:real_world}
\begin{tabular}{l c c c c c}
\toprule
& \multicolumn{4}{c}{Success Rate (Success / Total)} & \\
\cmidrule(lr){2-5}
Methods & Hang Up Tape & Transfer Tape & Put Cup Sleeve on Cup & Handoff Cup & Avg. Success Rate \\
\midrule
MoDE                  & 15/20 (75\%) & 11/20 (55\%)  & 12/20 (60\%)  & 0/20 (0\%)  & 47.50\% \\
\textbf{SMoDP (Ours)} & \textbf{19/20 (95\%)} & \textbf{18/20 (90\%)} & \textbf{16/20 (80\%)} & \textbf{20/20 (100\%)} & \textbf{91.25\%} \\
\bottomrule
\end{tabular}
\end{table*}

\subsection{Data Utilization Efficiency}
\label{sec:exp_data_eff}
We evaluate data utilization efficiency by training on only 10\%/20\%/50\% of \textsc{LIBERO-90} demonstrations, corresponding to 5/10/25 demonstrations per task.
This setting tests whether language-derived skill structure helps the policy reuse shared behaviors when task-specific demonstrations are scarce.
Figure~\ref{fig: data-util} summarizes model performances under different numbers of demonstrations. SMoDP consistently outperforms MoDE and MoDE (Pretrain) across all data regimes, while using fewer total parameters. 
Notably, SMoDP trained with 50\% of the demonstrations achieves performance comparable to MoDE (Pretrain) trained with the full target dataset, despite MoDE (Pretrain) using additional OXE pre-training and about $1.85\times$ more parameters.
With a comparable parameter count, SMoDP also shows better data efficiency than MoDE: using only 50\% of the training demonstrations, SMoDP outperforms MoDE trained on the full dataset. These results indicate that semantic skill supervision and skill-aware routing are particularly beneficial in low-data regimes, where purely data-driven routing is more likely to overfit task-specific correlations.



\subsection{Task Transfer and Compositional Generalization}
\label{sec:exp_transfer}
We evaluate whether SMoDP supports compositional transfer by reusing frozen experts and adapting only the skill predictor and router under few-shot target demonstrations. 
This setting tests whether previously learned experts can serve as reusable skill modules, while the lightweight skill predictor and router learn to recombine them for new task compositions.

\paragraph{\textsc{LIBERO-90} $\rightarrow$ \textsc{LIBERO-10}}
We first pre-train the model on \textsc{LIBERO-90} with 50 demos/task and then fine-tune it on \textsc{LIBERO-10} using 1/5/10 demos per task.
During fine-tuning, SMoDP freezes all expert parameters and updates only the skill predictor and router.
We compare against two baselines: MoDE+LoRA (fine-tuning attention weights with a matched trainable-parameter budget) and MoDE-scratch (training from scratch on \textsc{LIBERO-10}). 
Table~\ref{tab:libero_comparison} shows that SMoDP achieves the best performance across all three few-shot transfer settings, outperforming MoDE+LoRA by an average margin of 31.5 percentage points on LIBERO-10. Moreover, SMoDP updates only 13.7M parameters, accounting for 3.4\% of the full model. 
These results suggest that the source-suite experts capture reusable manipulation behaviors, and that new long-horizon tasks can be learned by adapting the skill prediction and routing modules rather than retraining the full policy.

\paragraph{\textsc{LIBERO-OBJECT/GOAL} $\rightarrow$ \textsc{LIBERO-GOAL-OOD}}
\textsc{LIBERO-GOAL-OOD} is constructed by recombining skill segments from \textsc{LIBERO-OBJECT} and \textsc{LIBERO-GOAL}, creating unseen verb--noun compositions at test time.
We fine-tune the model with 1/5/10 demonstrations per OOD task.
We use the same adaptation protocol as above: SMoDP updates only the skill predictor and router, while baselines employ MoDE+LoRA or MoDE-scratch.
The results are shown in Table~\ref{tab:libero_comparison}. SMoDP matches the strongest baseline in the 1-demo setting and achieves clear improvements in the 5-demo and 10-demo settings.
Unlike noise-conditioned routing in MoDE, SMoDP explicitly aligns routing decisions with language-grounded skill semantics, making adaptation focus on recombining reusable behaviors rather than relearning the entire action distribution from few demonstrations. This allows the skill predictor and router to adjust expert selection for new skill compositions, enabling parameter-efficient transfer to OOD task combinations.


\subsection{Real-world Results}
\label{sec:exp_real}
We further evaluate SMoDP in real-world multi-task learning settings on four bimanual manipulation tasks. Table~\ref{tab:real_world} reports the success rate over 20 rollouts per task. All methods are trained from scratch using the same demonstrations.
SMoDP consistently outperforms MoDE across all four tasks, improving the average success rate from $47.50\%$ to $91.25\%$.
On \emph{Hang Up Tape}, both methods achieve relatively high success rates, while SMoDP further improves performance from 75\% to 95\%. On \emph{Transfer Tape}, which involves picking up the tape, handing it over, and placing it on the table, SMoDP improves the success rate from 55\% to 90\%. On \emph{Put Cup Sleeve on Cup}, SMoDP improves from 60\% to 80\%, indicating more stable execution in object engagement and alignment. The largest gap appears on \emph{Handoff Cup}, where MoDE fails in all trials whereas SMoDP succeeds in all 20 rollouts.

We attribute these improvements to SMoDP's semantic skill decomposition and skill-aware routing, which enable knowledge sharing across related phases while preserving stable expert specialization.
Qualitative inspection further suggests that baseline failures often occur around high-variance transition points, such as grasping, handover, and object engagement.
By explicitly modeling latent skills and routing actions through semantically structured experts, SMoDP better disambiguates these critical phases and maintains stable execution in real-world multi-task settings.



\subsection{Ablation Studies}
\label{sec:exp_ablation}

\begin{table}[t]
    \centering
    \caption{Ablation Studies for SMoDP on \textsc{LIBERO-90}.} \label{tab:ablation}
    \begin{tabular}{l c}
    \toprule
    Variant & Avg. Success Rate $\pm$ Std \\
    \midrule
    \textbf{SMoDP (full)}         & \textbf{0.970 $\pm$ 0.002} \\
    w/o InterCL                    & 0.958 $\pm$ 0.002 \\
    w/o IntraCL                    & 0.957 $\pm$ 0.008 \\
    w/o InterCL \& IntraCL         & 0.946 $\pm$ 0.011 \\
    \midrule
    experts/layer = 3              & 0.957 $\pm$ 0.005 \\
    experts/layer = 6              & 0.958 $\pm$ 0.005 \\
    experts/layer = 8              & 0.950 $\pm$ 0.004 \\
    
    \bottomrule
    \end{tabular}
\end{table}

We validate the proposed framework by ablating its key components and architectural choices. Specifically, we study: (i) InterCL, which aligns predicted skill embeddings with language semantics; (ii) IntraCL, which encourages semantically consistent routing distributions; and (iii) the number of experts per MoE layer.
The ablation results are presented in Table~\ref{tab:ablation}. 

Removing either InterCL or IntraCL leads to a performance drop from $0.970$ to $0.958$ and $0.957$, respectively, while removing both further decreases the success rate to $0.946$. This indicates that the two objectives are complementary: InterCL improves language-grounded skill representation, whereas IntraCL directly regularizes routing behavior. 

For the number of experts, using 4 experts per layer achieves the best performance among the tested settings. Reducing the number of experts to 3 may limit specialization capacity, while increasing it to 6 or 8 does not further improve performance and may fragment the training signal across more experts. We therefore use 4 experts per layer as a trade-off between capacity, specialization, and parameter efficiency. With 6 MoE layers and top-2 routing, 4 experts per layer already provide sufficient routing diversity while keeping the model size comparable to MoDE.

\begin{figure}[t]
    \includegraphics[width=\linewidth]{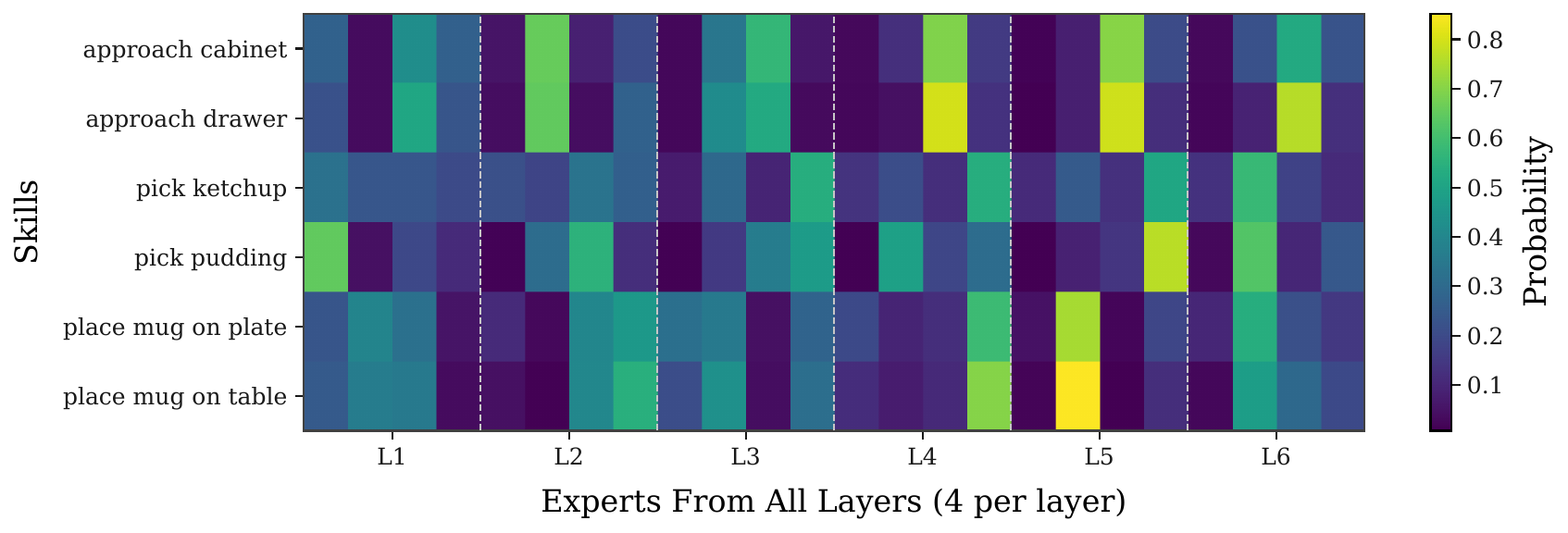}
    \caption{Expert activation heatmap for semantic skills across MoE layers. For each skill, we average expert activation probabilities over all occurrences of that skill during LIBERO-90 evaluation rollouts. Columns correspond to experts from layers 1--6, with 4 experts per layer.}
    \label{fig:expert_heatmap}
\end{figure}

We further visualize the learned routing behavior in Figure~\ref{fig:expert_heatmap}. For each semantic skill, we average the expert activation probabilities over all occurrences of that skill and display the activation pattern across all MoE layers. The heatmap shows that different skills induce distinct activation patterns, while semantically related skills exhibit similar expert usage. For example, related approach skills share more similar routing patterns than object-specific picking or placing skills. This provides qualitative evidence suggesting that SMoDP learns skill-dependent expert specialization beyond simple load balancing.

\subsection{Discussion and Limitations}
While SMoDP demonstrates significant improvements in semantic-aware multi-task learning, several limitations remain to be addressed in future work.

Coarse-grained Semantics: Encoding skills as verb-noun pairs provides a structured bottleneck but overlooks fine-grained control dynamics. For example, $\langle \texttt{pick up}, \texttt{cup} \rangle$ omits variations in speed, force, or trajectory, limiting the model's ability to adapt execution styles within the same high-level skill.

VLM-dependent Annotations: Offline annotation quality hinges on the VLM's zero-shot reasoning and priors. Coarse segmentation may miss subtle transitions, while excessive granularity can cause unstable routing. Visual challenges (e.g., motion blur, occlusion) further risk hallucinations, introducing noisy supervision via inaccurate boundaries or mislabeled skills.

Long-tail Skill Distribution: Robotic datasets often follow a long-tail distribution, where frequent primitives (e.g., reach, pick up) dominate over rare but critical skills (e.g., unscrew lid). This imbalance biases the router toward common skills, reducing expert specialization for infrequent actions.

\section{Conclusion} 
\label{sec:conclusion}

We introduce SMoDP, a semantically structured, skill-conditioned routing framework for diffusion policies that leverages language-based skill representations to guide expert selection, enabling more interpretable and structured expert specialization for multi-task robotic manipulation. Experiments on both simulation and real-world benchmarks demonstrate that SMoDP achieves superior data efficiency and parameter efficiency, while enabling compositional transfer through parameter-efficient fine-tuning.

\section*{Acknowledgments}
This work was supported by the Natural Science Foundation of China (62461160309), the NSFC-RGC Joint Research Scheme (N\_HKU705/24), and Hong Kong RGC (GRF 17201025, GRF 17200924).


\bibliographystyle{plainnat}
\bibliography{main}

\clearpage
\newpage
\appendix
\subsection{Details of Offline Semantic Skill Abstraction}
We use Qwen3-VL~\cite{bai2025qwen3vltechnicalreport} to automatically generate fine-grained skill annotations from task demonstrations without requiring a pre-defined skill set. The procedure has two stages: (i) skill discovery and initial temporal segmentation from the first demonstration of each task; and (ii) skill-aligned segmentation for the remaining demonstrations of the same task using the discovered skill descriptions as a fixed template.

For each task’s first demonstration, we use Qwen3-VL to segment the video into primitive skills with temporal boundaries and descriptions, using the following prompt:
\begin{lstlisting}
Task: {task_name}
Objects in the scene: {object_list_str}
Video duration: {time_length} seconds

Analyze the provided video and answer the following step by step:
1. Identify the primitive actions involved in the task (e.g., approach, pick up, place).
2. For each action, determine the temporal boundaries using video frame analysis:
    - Start time: When the action first becomes visible.
    - End time: When the action's completion is first verifiable.
    - Use 0.5-second intervals as the minimal time unit.
    - Ensure boundaries cover the entire {time_length}-second duration.
3. For each boundary, provide a concise description of the robot's action:
    - Omit the subject.
    - Use verb-noun structure (e.g., "approach object1", "place object2 on object3").
    - Each boundary should only contain one action.
    - Refer to objects using names from {object_list_str}.

Provide the final output in JSON format as follows:
<ANSWER> Explanation of the identified actions and their temporal boundaries. Do not use "and" to connect multiple actions in one boundary. </ANSWER>

{{
    "task_skill_count": <int>,
    "skill_details": [
        {{
            "skill_number": 1,
            "temporal_boundary": [<start_time>, <end_time>],
            "description": "<verb noun action description>"
        }},
        ...
    ]
}}
\end{lstlisting}


Within each task, we assume that all demonstration episodes have the same set of skills, each of which is in a unified verb-noun structure. After obtaining the skill descriptions and temporal boundaries for one demonstration episode, we use it as a template for cross-demonstration skill discovery and segmentation. 
Specifically, we prompt Qwen3-VL to perform temporal segmentation of raw episodes as follows:

\begin{lstlisting}
Task: {task_name}
Objects in the scene: {object_list_str}
Video duration: {time_length} seconds
Skill count: {skill_count}

Given a video of robot executing the task, it has been segmented into {skill_count} skills: {skill_description}.
Analyze the provided video and answer the following step by step:
1. Understand the skill descriptions and the video.
2. For each skill, determine the temporal boundaries using video frame analysis:
    - Start time: When the skill's action first becomes visible.
    - End time: When the skill's completion is first verifiable.
    - Use 0.5-second intervals as the minimal time unit.
    - Ensure boundaries cover the entire {time_length}-second duration.
    - Skill must be temporally contiguous and non-overlapping.
3. Keep each skill's description unchanged from the provided list.

Provide the final output in JSON format as follows:
<ANSWER> Explanation of the identified actions and their temporal boundaries. </ANSWER>
{{
    "task_skill_count": <int>,
    "skill_details": [
        {{
            "skill_number": 1,
            "temporal_boundary": [<start_time>, <end_time>],
            "description": "<original_skill_description>"
        }},
        ...
    ]
}}
\end{lstlisting}

\subsection{Additional Experimental Results}
All results in this section are from a single random seed and are provided as supplementary evidence; primary claims rely on the multi-seed results in the main paper.
\paragraph{Sensitivity to Annotation Noise}
To evaluate robustness to annotation quality, we perturb offline semantic supervision in two ways: (i) randomly replacing a fraction of semantic labels, and (ii) randomly shifting a fraction of temporal boundaries by $\pm 10$ steps. The results are shown in Table~\ref{tab:sup_noise}.
SMoDP remains stable under moderate corruption, with only small performance drops up to 30\% noise. We attribute this behavior to the fact that semantic labels are used as auxiliary supervision for skill prediction and routing regularization, rather than direct action targets.
When semantic label noise reaches 50\%, performance drops to 95.1\%, indicating that severe corruption starts to mislead expert specialization.

\begin{table}[t]
\centering
\caption{Sensitivity analysis on LIBERO-90 under different noise ratios.}
\label{tab:sup_noise}
\vspace{-0.2cm}
\begin{tabular}{c c c c c}
\toprule
Noise Ratio & 0\% & 10\% & 30\% & 50\%\\
\midrule
Semantic Label & 96.9\% & 96.3\%  & 96.1\% & 95.1\% \\
Temporal Boundary& 96.9\% & 97.2\%  & 96.7\% & 96.1\% \\
\bottomrule
\end{tabular}
\vspace{-0.2cm}
\end{table}

\paragraph{Comparison with Auxiliary Routing Signal}
To isolate whether improvements come from semantics or merely from adding an auxiliary routing signal, we replace language-based skill descriptions with phase IDs while keeping all other settings unchanged. Table~\ref{tab:sup_signal} shows that semantic labels consistently outperform phase IDs across all data regimes (5/10/25/50 demos), with larger margins in lower-data settings.
This suggests that language-grounded semantics provide stronger structure for expert reuse than purely index-based phase supervision.

\begin{table}[t]
\centering
\caption{Routing signal comparisons on LIBERO-90 under different numbers of demonstrations for training.}
\label{tab:sup_signal}
\vspace{-0.2cm}
\begin{tabular}{c c c c c}
\toprule
Demos & 5 & 10 & 25 & 50\\
\midrule
Semantic Label & 79.9\% & 86.7\%  & 94.7\% & 96.9\% \\
Phase ID & 76.2\% & 83.0\%  & 93.5\% & 94.7\% \\
\bottomrule
\end{tabular}
\vspace{-0.2cm}
\end{table}



\begin{figure}[t]
    \centering
    \includegraphics[width=\linewidth]{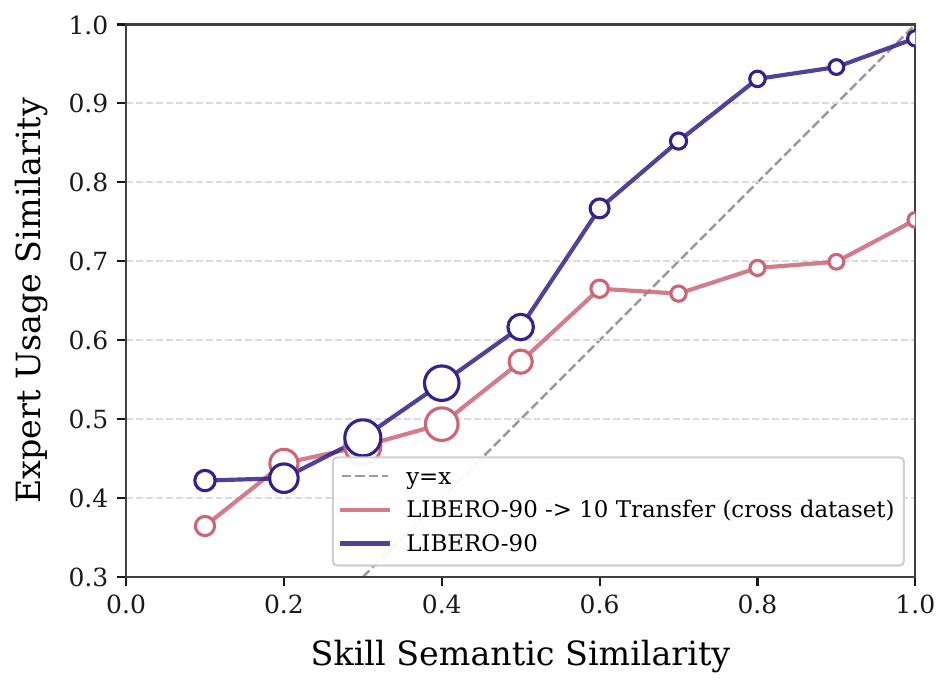}
    \caption{Correlation between skill semantics and expert usage.}
    \label{fig:sup_skill_expert_corr}
\end{figure}

\paragraph{Skill Semantics and Expert Reuse Correlation}
We further examine whether semantically related skill phases share similar expert-routing patterns.
Figure~\ref{fig:sup_skill_expert_corr} reports the relationship between skill semantic similarity and expert-usage similarity under two settings.
First, we compute this correlation within LIBERO-90 evaluation rollouts by measuring, for each pair of skill phases, the cosine similarity between their semantic representations and the corresponding cosine similarity between their expert-usage vectors.
Second, we evaluate the same relationship across transfer by comparing skill phases from the transferred LIBERO-10 benchmark with rollout data from the original LIBERO-90 benchmark.
In both settings, we observe a positive trend: skill phases with higher semantic similarity tend to induce more similar expert-usage patterns.
This indicates that the learned routing behavior is aligned with the semantic structure of skills, rather than being solely determined by instance-specific trajectory statistics.
Overall, this analysis provides additional evidence that SMoDP organizes experts according to reusable semantic skill structure, and that such structure is partially preserved after transfer to new tasks.

\begin{figure*}[t]
    \centering
    \includegraphics[width=\linewidth]{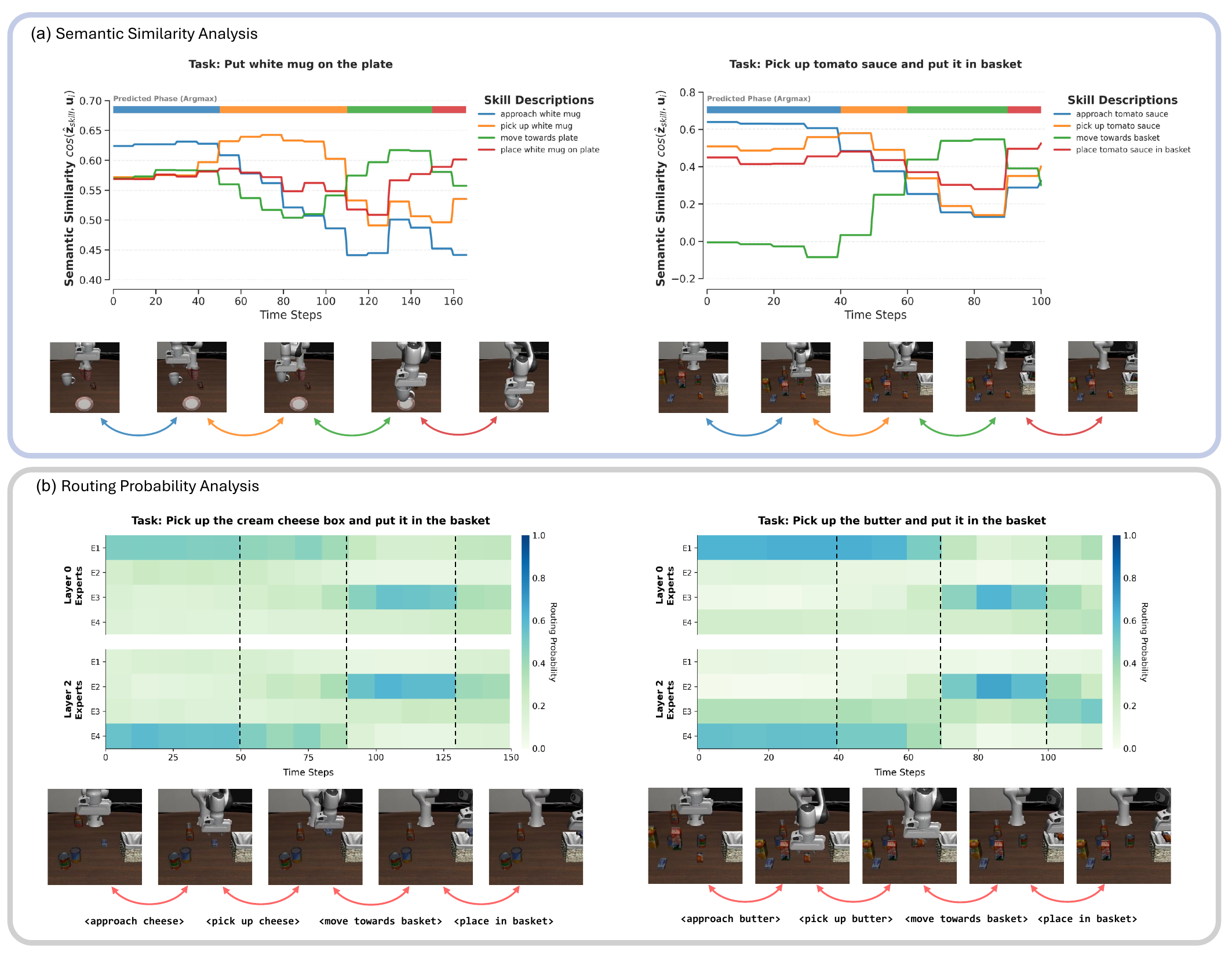} 
    \caption{Visualization of SMoDP: (a) \textbf{Semantic Similarity Analysis}: We visualize the skill token, the output of the skill predictor, by computing its cosine similarity with all skill annotation representations for that task.  (b) \textbf{Routing Probability Analysis}: We visualize the routing probabilities, the outputs of the router, as heatmaps. Each column corresponds to a time step and each row to an expert; color intensity indicates the activation probability. Dashed vertical lines mark the manually annotated phase boundaries based on routing patterns (approach, pick up, move towards basket, place in basket). Our model has six MoE layers, and we visualize the router outputs for two representative layers (layers 0 and 2) for clarity, as the remaining layers exhibit similar patterns.}
    \label{fig:vis}
\end{figure*}

\subsection{Visualization on Skill Prediction and Policy Routing}

We visualize some examples to qualitatively demonstrate the interpretability and routing consistency of our approach, which is shown in Figure~\ref{fig:vis}.

In Figure~\ref{fig:vis}(a), we present a semantic similarity analysis for two representative tasks: (i) \emph{Put white mug on the plate} and (ii) \emph{Pick up tomato sauce and put it in basket}. For each time step, we compute the cosine similarity between the predicted skill token and all skill annotation embeddings for that task. The colored curves show how the similarity to each skill evolves over time, while the bar on the top indicates the predicted skill whose semantic similarity dominates others. We observe that the skill with dominant similarity largely aligns with the executing primitives of the manipulation sequence. For example, in the mug task, the similarity to the approach-mug skill is highest at the beginning of the trajectory, then gradually gives way to pick-up-mug, followed by move-to-plate and finally place-on-plate. A similar progression is observed for the tomato sauce task. The transitions between skills are smooth rather than abrupt, with multiple skills exhibiting comparable similarity near the phase boundaries. This soft switching behavior aligns with the fact that real trajectories often contain short periods of overlap between consecutive sub-tasks (e.g., finishing approach while initiating grasp), and provides an interpretable view of how the model composes skills over time.

Figure~\ref{fig:vis}(b) further analyzes the routing behavior of the Mixture-of-Experts (MoE) policy. We plot the routing probabilities produced by the router as heatmaps for two MoE layers (layer 0 and layer 2) on two different tasks: (i) \emph{Pick up the cream cheese box and put it in the basket} and (ii) \emph{Pick up the butter and put it in the basket}. Despite differences in object geometry and scene configuration, similar experts are consistently activated during semantically corresponding phases of the two tasks. This cross-task consistency suggests that the experts specialize in reusable motor primitives rather than overfitting to specific instances, which supports our design goal of skill modularity.

Taken together, these visualizations indicate that (i) the learned skill tokens align with human-interpretable task phases, and (ii) the routing mechanism exhibits structured, repeatable patterns across tasks. These results provide qualitative evidence that SMoDP decomposes multi-stage manipulation into meaningful skills and routes observations to specialized experts in a consistent and interpretable manner.

\end{document}